\pdfoutput=1

\documentclass[11pt]{article}

\usepackage[
]{acl}
\usepackage[bottom]{footmisc}
\usepackage{times}
\usepackage{latexsym}
\usepackage{amsmath}
\usepackage{amsthm}
\usepackage{amsfonts}
\usepackage{amssymb}
\usepackage{enumitem}
\usepackage{multirow}
\usepackage{geometry}
\usepackage{booktabs}
\usepackage{tcolorbox}
\usepackage[T1]{fontenc}

\usepackage[utf8]{inputenc}

\usepackage{microtype}

\usepackage{inconsolata}

\usepackage{graphicx}

\title{Identity-Driven Hierarchical Role-Playing Agents }

\author{
 \textbf{Libo Sun\textsuperscript{1}},
 \textbf{Siyuan Wang\textsuperscript{2}},
 \textbf{Xuanjing Huang\textsuperscript{3}},
 \textbf{Zhongyu Wei\textsuperscript{1,4}},
\\
\\
 \textsuperscript{1}School of Data Science, Fudan University, China, \\
 \textsuperscript{2}University of Southern California, \\
 \textsuperscript{3}School of Computer Science, Fudan University, China, \\
 \textsuperscript{4}Research Institute of Intelligent Complex Systems, Fudan University, China
\\
 \texttt{lbsun23@m.fudan.edu.cn}
 \texttt{siyuanwang1997@gmail.com} \\
 \texttt{\{xjhuang, zywei\}@fudan.edu.cn} \\
}

\begin{document}
\maketitle
\begin{abstract}

Utilizing large language models (LLMs) to achieve role-playing has gained great attention recently. The primary implementation methods include leveraging refined prompts and fine-tuning on role-specific datasets. However, these methods suffer from insufficient precision and limited flexibility respectively. To achieve a balance between flexibility and precision, we construct a Hierarchical Identity Role-Playing Framework (HIRPF) based on identity theory, constructing complex characters using multiple identity combinations. We develop an identity dialogue dataset for this framework and propose an evaluation benchmark including scale evaluation and open situation evaluation. Empirical results indicate the remarkable efficacy of our framework in modeling identity-level role simulation, and reveal its potential for application in social simulation. The source code is available in \url{https://anonymous.4open.science/r/HIS-7DBB}.

\end{abstract}

\section{Introduction}


The advanced natural language generation capabilities of Large Language Models (LLMs) enable the creation of convincing human simulacra, which have attracted significant research interest\cite{wang2023rolellm, shao2023character, tu2024charactereval}. 
By building Role-Playing Agents (\textbf{RPAs}) \cite{li2023camel},  LLMs can emulate various personas, ranging from specific individuals to typical representatives of demographic groups. 
The applications of RPAs offer benefits beyond entertainment \footnote{\url{https://character.ai/}} and have significant influences in fields such as psychotherapy\cite{stade2024large}, economics\cite{fu2023improving} and social studies\cite{grossmann2023ai}.


\begin{figure}[h]
    \centering
    \includegraphics[width=0.57\textwidth]{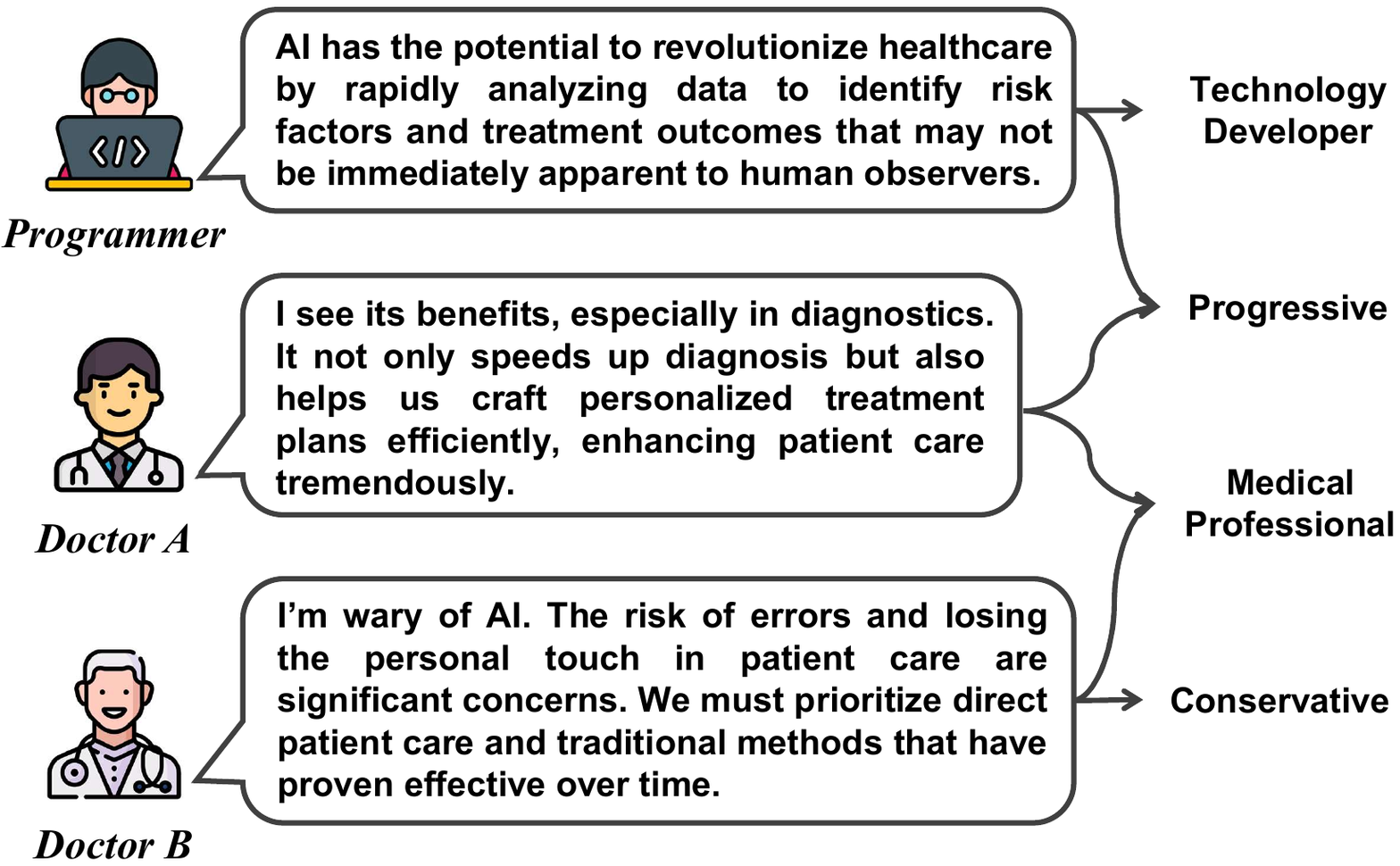}
    \caption{Individuals possess various identities that shape their unique characteristics and interactions. Shared identities often foster commonalities, while differing identities highlight diversity.}
    \label{fig:intro}
\end{figure}

The implementation of Role Playing Agents can be categorized into two primary approaches: prompt-based strategies and fine-tuning methods. Prompt-based strategies employ detailed instructions to direct LLMs to adhere to specific expression and behavior patterns of roles\cite{park2023generative}, offering advantages in terms of convenience and flexibility. However, they often do not capture the intricacies of complex personas with high fidelity. In contrast, fine-tuning methods achieve more precise simulations of complex roles by utilizing specialized data collection and training processes\cite{shao2023character,zhou2023characterglm}. As the price, these methods are costly and exhibit less flexibility due to the requirement for role-specific data processing and training.


Identity Theory \cite{stryker1968identity} in sociology argues that an individual's self is composed of multiple identities such as familial identities (e.g., spouse, parent) and occupational identities (e.g., artist, doctor). These identities shape behavior by carrying specific meanings and expectations that guide how individuals act\cite{burke1981link}. As illustrated in Figure \ref{fig:intro}, Programmer and Doctor A, both holding progressive identities, are optimistic about AI's potential benefits in the medical field. Conversely, Doctor B, possessing a conservative stance, emphasizes the risks and adverse impacts of AI on medical treatments. While Doctors A and B evaluate AI from a medical standpoint as medical professionals, the Programmer primarily considers its technical aspects since his technological background. In summary, roles possess both shared and distinct identities. Shared identities encourage commonality, while distinct identities emphasize diversity. This inspires us to implement role-playing from the identity level, using combined identities to form complex roles. However, this approach faced with two main challenges: 1) organizing and accurately modeling diverse identities in the system, and 2) ensuring controllable and flexible integration of these identities.



To address these challenges, we introduce the Hierarchical Identity Role-Playing Framework (HIRPF) which constructs and integrates identities from diverse categories to offer a versatile role-playing solution. We focus on two dimensions of identities: personality traits and professions, designed identity isolation and explict control to address the two challenges. 
Identity isolation includes intra-level and inter-level isolation.
Inspired by the methods \cite{liu2023moelora, wu2023mole} that combine the Lora\cite{hu2021lora} and MoE\cite{shazeer2017outrageously}, we specify the corresponding Lora parameters for each identity in each dimension to achieve intra-level isolation. Furthermore, we alternately insert identities of different dimensions into the model blocks to achieve inter-level isolation. Regarding how to achieve controllable and flexible integration between different identities, we achieve explicit control with hard masking and soft routing. The hard mask is constructed according to the input identities to be activated and fed into the soft routers, so that the fusion weight of the soft routers output of the inactive identity module is ignored, enabling flexible integration of only interested identities. To construct a dataset for this framework, we utilize ChatGPT to initially generate identity dialogue datasets encompassing both singular and multiple identities. We fine-tune Llama2-chat on this dataset to develop a model for identity-level role-playing. In order to evaluate the role-playing ability of identity-level agents, we propose a systematic benchmark, including scale tests and open-ended situation tests. Our model's performance on this benchmark demonstrates better identity-level fitting capabilities and has potential for application in social simulation.

In general, our contributions are three-fold:

\begin{itemize}
    \item We propose the Hierarchical Identity-Level Role-Playing (HILRP) Framework, a novel framework that constructs roles from the identity level. This framework strikes a balance between flexibility and fidelity and allows diverse variations.
    \item We introduce a specialized identity dialogue dataset for our framework, including with 20,685 multi-turn role-playing dialogues for singular and multiple identities, involving personality traits and professions.
    \item We develop Identity-Eval, a systemic benchmark for identity-level role-playing evaluation which includes scale tests and open-end situation tests. This benchmark is used to measure the agent's fidelity to single identity and its ability to integrate multiple identities.

\end{itemize}

\section{Methodology}

\subsection{Hierarchical Identity Role-Playing Framework}


Identity Theory asserts that individuals have multiple identities stemming from their affiliations with various social groups, categorized by criteria such as occupation, ethnicity, and gender. 
The relationship between identities is either exclusive or compatible. Identities within the same category are generally mutually exclusive (e.g. progressive v.s conservative), while those from different categories may coexist within an individual (e.g. a progressive artist). To this end, we devise the Hierarchical Identity Role-Playing Framework (HIRPF) to construct identity-level role-playing from personality traits and professions, which is distinguished by two features: identity isolation and explicit control.


\begin{figure*}[t]
    \centering
    \includegraphics[width=0.85\textwidth]{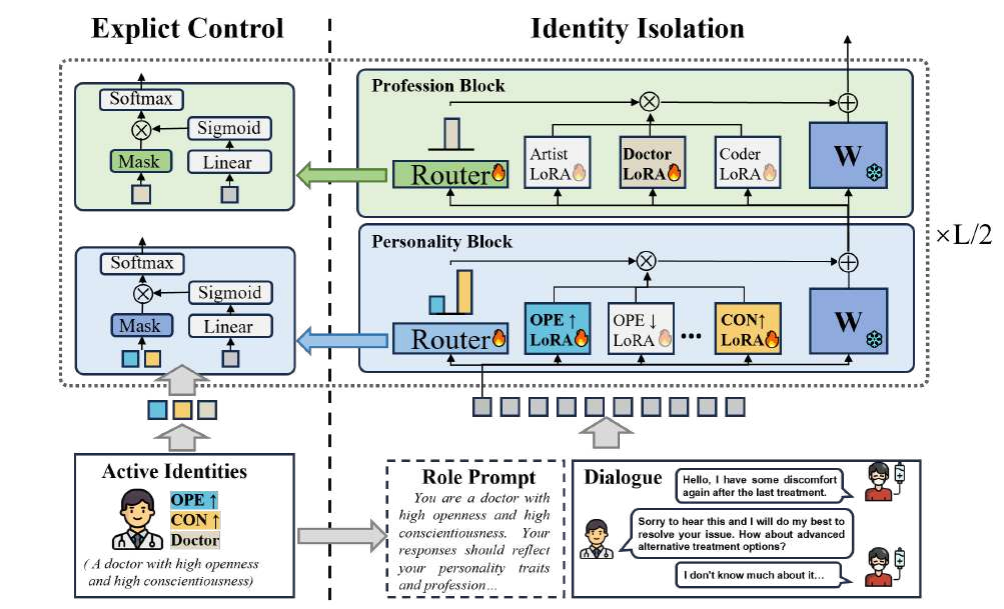}
    \caption{Structure of Hierarchical Identity Role-Playing Framework (HIRPF). HIRPF ensures identity isolation via intra-level and inter-level separation and uses control routers in each block to update only activated identities during training, maintaining distinct identity training. The figure shows an example of the fine-tuning process, where $L$ represents the total number of blocks of the backbone model.
}
    \label{fig:enter-label}
\end{figure*}

\paragraph{Identity Isolation} Identity isolation is used to model mutual exclusivity between identities, which operates on two levels: intra-level and inter-level isolation. For intra-level, refer to \cite{wu2023mole} and assume that $W_0 \in \mathbb{R}^{m\times d}$ represents the parameter matrix of the pre-trained LLM, we specify corresponding Lora matrices $B_{i}\in\mathbb{R}^{m\times r}$ and $A_{i}\in\mathbb{R}^{r\times d}$ for each identity in the same category (personality or profession), where $i$ indicates the $i$-th identity in the category and $r$ indicates the rank of Lora. For inter-level, Lora matrices for each category are inserted alternately into blocks in backbone models, which distinguish different categories of identities while reducing the number of parameters.

\paragraph{Explicit Control} To model compatibility between identities, allowing flexible and controlable fusion of identities, we achieves explicit control. This enables users to tailor the identity composition for their requirements. Given the identities to be activated $\{i_{0}, i_{1}, ..., i_{a} \}$ in $j$-th block, where $a$ indicates the number of active identities, the updated attention parameters $W_{j}$ are derived as follows:

\[
\begin{aligned}
    W_{j} &= W_{0} + \sum_{k=1}^{n}w_{jk} \cdot B_{k} A_{k} \\
    w_{j} &= \text{Gate}(e_{j}) \\
    &= \text{Softmax}(M_{j} \cdot \sigma(e_{j})) \in \mathbb{R}^{n}
\end{aligned}
\]

\noindent where $n$ indicates the total number of identities in $j$-th block, $e_{j} \in \mathbb{R}^{n}$ represents the input features for the block, $\sigma$ is the sigmoid function that scales $e_{j}$ into a positive range, and $M_{j} \in \mathbb{R}^{n\times n}$ is a diagonal mask matrix. In $M_{j}$, diagonal elements corresponding to activated identities are set to 1 while others are set to $-\infty$, which ensures that only the weights for the active identities remain positive and influence the model's output, maintaining clear control over the model's identity combination.

\paragraph{Training \& Inference} During the training phase, HIRPF receives dialogues and associated active identities. Role prompts are initially constructed from these identities, combined with the dialogue, and tokenized before model input. During forward propagation, routers apply explicit control based on active identities, updating only related routers and Lora parameters while keeping other parameters unchanged. In the inference phase, users activate specific identities and role prompts to start conversations.


\subsection{Identity Dialogue Dataset}
\label{sec:dataset}

Most existing role-playing datasets focus on the role level \cite{shao2023character,zhou2023characterglm} without providing identity-level data. While some datasets cater to specific personalities\cite{zhang2018personalizing}, they fall short in covering interactions across multiple identity categories. Consequently, we have developed an identity dialogue dataset that includes dialogues featuring individual personality traits, professions, and multiple identities.

\paragraph{Single Personality}

The construction of dialogues with singular personality traits follows the pipeline outlined in \cite{shao2023character}: occasion generation, plot generation, and dialogue generation. Initially, we employ ChatGPT \footnote{We use gpt-3.5-turbo-1106 throughout this paper.} to create a variety of daily occasions tailored to each personality factor, aiming to provoke varied behavioral responses based on personality polarities. Subsequently, we generate corresponding plots based on these daily occasions and specified personality traits and tendencies, which serves to delineate the interlocutors and the topics of their interactions. Finally, we generate specific dialogues based on the developed plots. Refer to Appendix \ref{sec:appendix} for detailed prompts.

\paragraph{Single Profession}

The construction of dialogues with specific professions involves two phases: topic generation and dialogue generation. In topic generation, we use ChatGPT to create a series of topics related to various occupations that require professional knowledge. During dialogue generation, we select from a pool of occupations and topics, requiring individuals to demonstrate extensive knowledge in their field while appearing uninformed or evasive on unrelated topics. This approach reinforces the distinctions between professional identities, preventing out-of-character responses. For detailed prompts, see Appendix \ref{sec:appendix}.

\begin{table}[b]
\centering

\begin{tabular}{|l|c|}
\hline
\# of samples & 20,685 \\
\hline
Avg. \# of turns & 9.52 \\
\hline
Avg. \# of words per response & 11.59 \\
\hline
Avg. \# of words per dialogue & 220.67 \\
\hline
Avg. \# of active identities & 1.67 \\
\hline
\end{tabular}
\caption{Statistics for identity dialogue dataset.}
\label{dataset}
\end{table}

\paragraph{Multiple Identities \& Re-annotation}

It has been observed that using ChatGPT to simulate a character with multiple identities and expressing all of them within a single, constrained dialogue poses significant challenges while striving to maintain the naturalness of the conversation. Concurrently, verifying whether the dialogues generated by ChatGPT accurately reflect the requested identities necessitates further examination. Consequently, we have restricted our construction to dialogues representing pairwise combinations of personality traits and professional identities, emulating the process used for constructing dialogues for individual characters. Additionally, all dialogues generated have been re-annotated, requiring ChatGPT to identify the characters' identities based solely on the dialogues. For detailed prompts, refer to Appendix~\ref{sec:appendix}.
The statistical results of the data set are shown in Table~\ref{dataset}.

\section{Benchmark}

To evaluate identity-level agents, we construct a benchmark including scale tests and open-ended situation tests.

\subsection{Scale Test}

We first conducted scale tests to assess the fit of each identity, including personality scale and profession scal.

\paragraph{Personality Scale}

We use BF-marker-1003\footnote{\url{https://ipip.ori.org/newBigFive5broadKey.htm}} as the personality scale, which is one of the most widely used scales in Big-Five. In BFM, there are 20 items in each trait. Each item is a statement that reflects the positive or negative aspect of one trait. The scale is originally a five-dimensional Likert scale, and subjects need to choose according to the item how much it is consistent with them.

\paragraph{Profession Scale}
The profession scale aims to evaluate the character's suitability for a specific profession. Since no similar scales available, we model the personality scale and construct a profession scale. In the profession scale, each item for one occupation is a statement related to the professional knowledge or skills. It is used to measure whether the character imitates the relevant knowledge of the corresponding profession, and at the same time, the character is expected to be ignorant of non-professional knowledge. There are 20 items for each occupation in this scale.

\begin{figure*}[t]
    \centering
    \includegraphics[width=\textwidth]{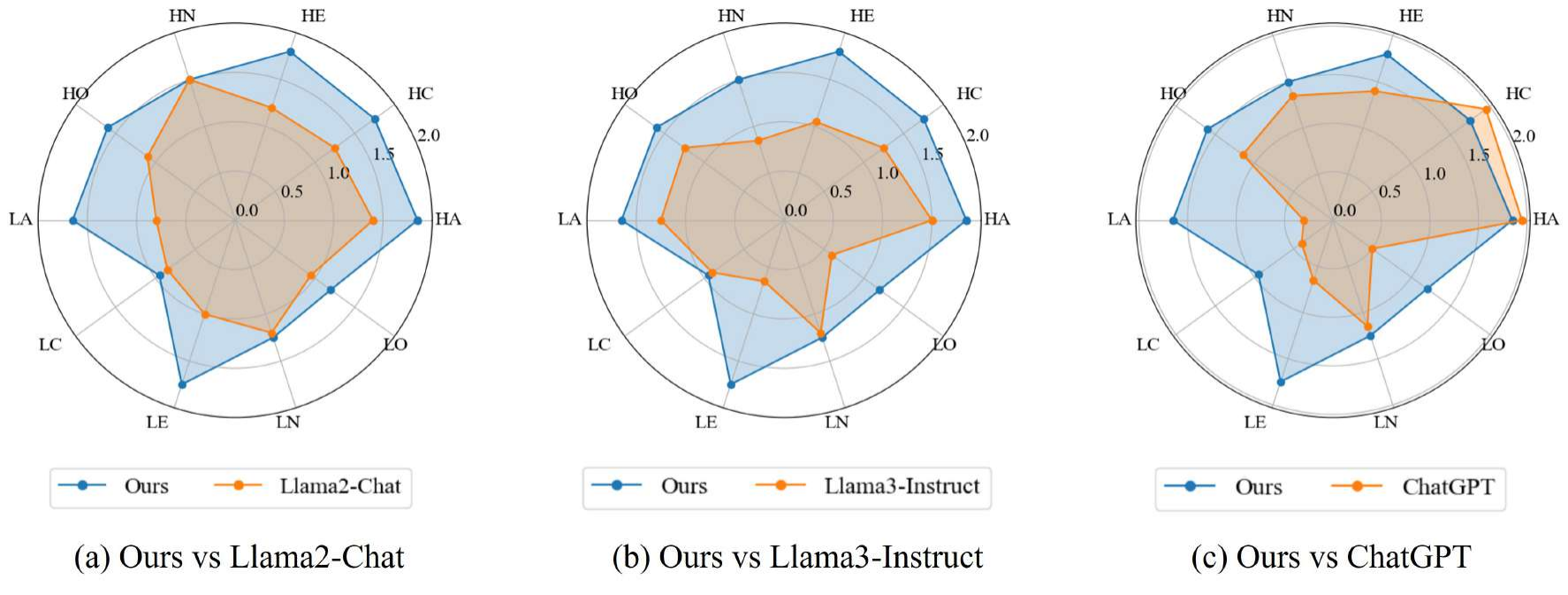}
    \caption{Scale results for single trait simulations. Compared with models with the same magnitude number of parameters using prompt strategies, our method can achieve more obvious simulation of personality trait tendencies. Compared with ChatGPT, our method has a more obvious fitting effect on most traits, especially on negative traits (low agreeableness, low conscientiousness, low extraversion, etc.).
}
    \label{fig:trait_scale}
\end{figure*}

\paragraph{Test Implementation}

Although the original personality scale was a five-dimensional Likert scale, issues often arose during implementation where the agent's outputs did not align with its explanations. To address this, we introduced an interactive scale measurement method. This method involves using ChatGPT to question the agent and engage in multiple rounds of dialogue to elicit detailed responses about each item. ChatGPT then evaluates these responses to determine which scale option the agent matches, using self-consistency checks to finalize the decision. We have established three rounds of interaction and 5 repetitions of self-consistency in the evaluation phase.


\subsection{Open-end Situational Test}

In addition to the fit of a single identity, we also want to be able to evaluate the accuracy of multi-identity fusion. Rather than each item on a scale targeting a single personality trait or profession, there is more information by observing a character's reaction in a specific situation. To this end, we built open-ended situational tests.

\paragraph{Situation Design} We use ChatGPT to create eight distinct scenarios, each corresponding to a specific trait and profession. These scenarios include background information and interactive character settings. In each scenario, the agent interacts with characters driven by ChatGPT based on the background. Unlike the interactive scale test, which focuses on specific factors, our situational test centers around a dialogue within a given situation, allowing for the expression of various character identities. Moreover, the characters in the situational test display emotional tendencies, differing from the neutral interviewer in scale tests and enhancing the realism of the interactions.



\paragraph{Test Implementation} 
We test 971 combinations of personality traits and professions across 8 scenarios, with each dialogue limited to 4 rounds. Initially, the interactive character initiates the conversation using background and role settings, followed by interactions with the agent. Post-dialogue, ChatGPT evaluates the dialogues to identify the displayed personality traits and professions of the agents. Each evaluation is repeated 5 times for consistency in the final judgment. The agent must remain anonymous during the conversation and any breach is recorded as an error to maintain result comparability. Agents can express unspecified identities without affecting the final accuracy.


\begin{figure*}[t]
    \centering
    \includegraphics[width=\textwidth]{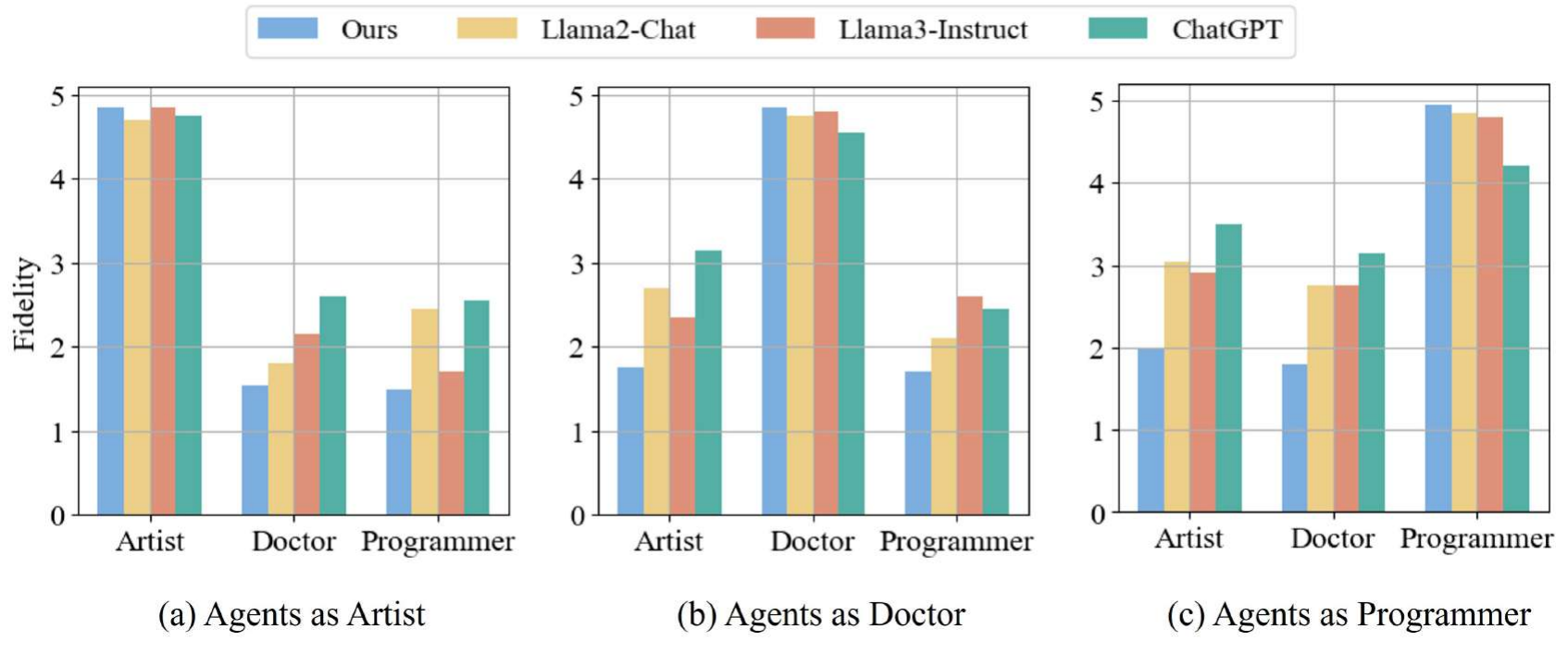}
    \caption{Scale results for single profession simulation. Compared to other prompt-based models, our model's professional agents perform best in their corresponding profession and score lower on unrelated professional field questions. This demonstrates our method's fidelity in simulating specific professions while clearly distinguishing between different professional identities. 
}
    \label{fig:prof_scale}
\end{figure*}

\section{Experiments}

\subsection{Experimental Setup}

\paragraph{Implementation Details}
We apply our framework HIRPF on top of Llama2-7b-chat and fine-tune it using our identity dialogue dataset. The $\alpha$ and $r$ of the LoRA method are both set to 16. There are 10 personality identities (positive and negative for each trait) and 3 professional identities (artist, doctor, and programmer). We set the learning rate to $1\times10^{-4}$, batch size to 8, gradient accumulation step to 4, employ
AdamW as the optimizer and train the model 5 epochs on 8 NVIDIA RTX4090 GPUs.

\paragraph{Models for comparision}

Since we are the first to propose a role-playing method from the identity level, there is currently no equivalent fine-tuning model to compare with us, so we mainly compare with prompt-based methods, and the comparison with role-specific fine-tuning methods will be left in the discussion part \ref{sec:discussion}. We selected the following models for comparison:

\begin{itemize}
    \item LLaMA-2\cite{touvron2023llama}: Developed by Meta, LLaMA-2 is an open-source language model. For comparative analysis, we have chosen the LLaMA-2-7b-chat variant.
    \item LLaMA-3: The latest in the LLaMA series, LLaMA-3 represents an advancement over its predecessors with several improvements. We have selected the LLaMA-3-8B-Instruct model for comparison.
    \item ChatGPT: Optimized for dialogue, ChatGPT excels in conducting conversations in a natural manner. In our experiments, we employ the gpt-3.5-turbo-1106 version.
\end{itemize}

\noindent All models use this prompt template \ref{sec:appendix} to build system prompts for the corresponding characters.

\subsection{Main Results}

\paragraph{Single Trait Simulation} 

Figure~\ref{fig:trait_scale} illustrates a comparison between our method and baseline models in accurately representing single personality polarities. The original scores on the scale range from 1 to 5, with higher scores indicating more pronounced positive traits and lower scores corresponding to negative traits. The scores shown in the figure represent the absolute differences from the neutral score of 3, The higher the value, the higher the degree of fitting to the trait in the corresponding direction. Our method shows superior performance across all dimensions compared to models with similar parameter sizes, including Llama2-7B-chat and Llama3-8B-Instruct. Additionally, our method surpasses the ChatGPT in depicting most personality traits, especially negative ones such as low agreeableness, low conscientiousness, low extraversion, and low openness. This advantage is largely due to the limitations of ChatGPT's ethical and moral frameworks, which hinder its ability to accurately express negative traits and realistically simulate behaviors associated with these traits.

\paragraph{Single Profession Simulation}

Figure~\ref{fig:prof_scale} displays the performance of our method compared to baseline models using the professional scale. In the figure, the horizontal axis represents the various occupational dimensions covered by the scale, while the vertical axis quantifies the fidelity of representation across specific occupations. Agents developed through our method consistently achieve the highest fidelity scores in the dimensions consistent with their professions and markedly lower scores in non-professional dimensions. This distinction is critical as it shows that our method not only excels in accurately fitting into specific occupational roles but also effectively discriminates between different professional identities. This precision in role-fitting and identity distinction is especially valuable in applications requiring accurate role portrayal and differentiation, such as training simulations, automated customer service, and interactive educational systems.

\begin{table*}[h]
\centering
\begin{tabular}{@{}lcccccc@{}}
\toprule
Model           & AGR   & CON   & EXT   & EMS   & OPE   & Profession \\
\midrule
ChatGPT         & 52.95 & \textbf{50.71} & \textbf{60.38} & \textbf{47.57} & \textbf{44.81} & \textbf{19.34} \\
Llama3-8B-Instruct & \underline{55.69} & 48.78 & 40.89 & 39.6  & 23.46 & 9.02 \\
Llama2-7b-Chat  & 50.25 & 31.02 & 40.51 & 30.32 & 20.97 & 4.78 \\
\hline
Ours     & \textbf{60.59} & \underline{48.89} & \underline{41.18} & \underline{43.29} & \underline{27.6}  & \underline{14.49} \\
\bottomrule
\end{tabular}
\caption{Results from the Open Situation Test. The model is given specific identities and interacts anonymously within set scenarios. ChatGPT then identifies the agent from the dialogue, checking if it matches the assigned identities. Our model excels in agreeableness and ranks second to GPT in other categories. Notably, it shows significant improvement in fitting professional identities over methods that rely only on prompts.
}
\label{tab:open_test}
\end{table*}

\paragraph{Open-end Situation Test}

Table~\ref{tab:open_test} displays the outcomes of an open-ended situational test designed to assess whether an agent accurately represents the required identity in performance. Each column within the table measures the precision with which the agent's generated dialogues align with a predefined identity, assessing how well the agent captures the essence of the specified character traits. The results reveal that our method achieves superior accuracy in simulating the trait of agreeableness, outperforming the ChatGPT in this dimension. Although our method is second only to GPT in other personality dimensions, it shows a notable improvement in recognizing occupational traits compared to strategies that rely solely on prompts. This enhancement stems primarily from our approach's focus on establishing clear professional boundaries.

\begin{figure}[t]
    \centering
    \includegraphics[width=0.5\textwidth]{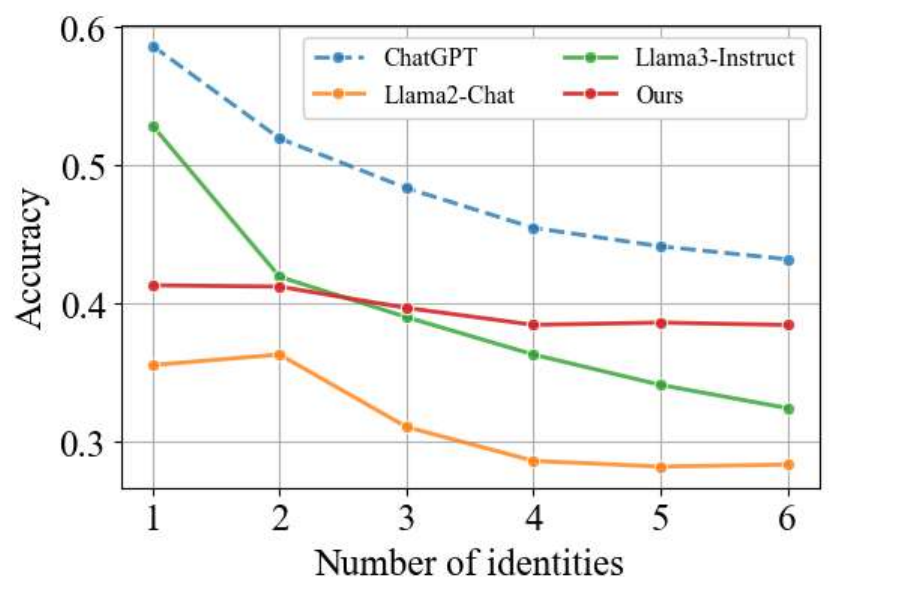}
    \caption{Identification accuracy declines  as the number of assigned identities increases. Despite this trend, our model shows a more gradual decrease and maintains the second-highest accuracy behind ChatGPT for more than two identities. This indicates our model's effective capability in integrating multiple identities.}
    \label{fig:acc_plot}
\end{figure}

Figure~\ref{fig:acc_plot} illustrates how the accuracy of identity simulation changes as the number of required identities increases. All methods show a noticeable drop in accuracy with an increase in the number of identities. Our method experiences a gradual decline in accuracy, ranking just behind ChatGPT when simulating more than two identities. This indicates that our method effectively handles the integration of multiple identities. Particularly in the professional dimension, our method significantly improves the fitting accuracy compared to methods that only use prompts.

\paragraph{Ablation Study}

To validate the efficacy of identity isolation and explicit control within the HIS framework, we developed two alternative models for ablation studies:

\begin{itemize}[itemsep=1pt]
    \item Llama2-Identity: This model involves directly fine-tuning the Llama2-7b-chat using the traditional Lora method with the identity dialogue dataset.
    \item HIRP-Dense: In this variant, all identity modules are activated for each inference, employing only role prompts to facilitate identity fusion.
\end{itemize}

The ablated models were evaluated through scale tests and open-end situation assessments, where a decline in various performance metrics was observed. These results substantiate the contribution of the two key features in our proposed framework. Detailed findings are documented in the Appendix~\ref{sec:appendix}.

\section{Discussion}
\label{sec:discussion}
\subsection{Comparision with Role-specific model}

The framework in this study is not compared with role-specific models as it emphasizes demographic effects in broad social simulations rather than detailed role simulations. It focuses on group dynamics over individual identities to 
better represent general characteristics of demographic groups. This approach is particularly relevant for social governance and group decision-making. We will demonstrate the applications of our framework in social simulations to highlight its potential in Section~\ref{social_simulation}.

\subsection{Social Simulatin Application}
\label{social_simulation}
\paragraph{Questionnaire}

Questionnaires are crucial in sociology for understanding diverse population responses to social issues. Researchers can leverage LLMs to simulate various demographic groups  to predict how different segments might react to policies or societal changes.

\begin{figure}
    \centering
    \includegraphics[width=0.48\textwidth]{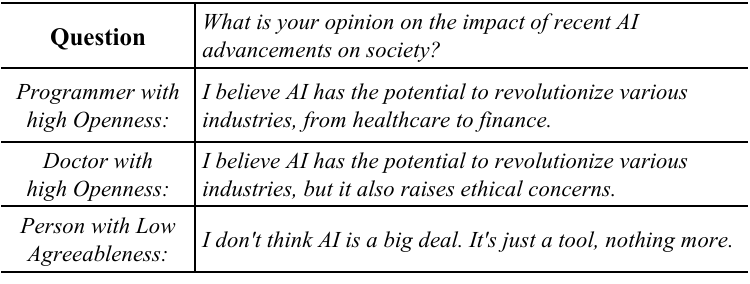}
    \label{fig:enter-label}
    \caption{An example for questionnaire with our framework.}
\end{figure}

\paragraph{Debate between Differentiated Individuals}
AI simulations of group interactions reveal insights into differences, enable testing of communication strategies, and provide a safe environment for exploring conflict resolution without real-world impacts.

\begin{figure}
    \centering
    \includegraphics[width=0.48\textwidth]{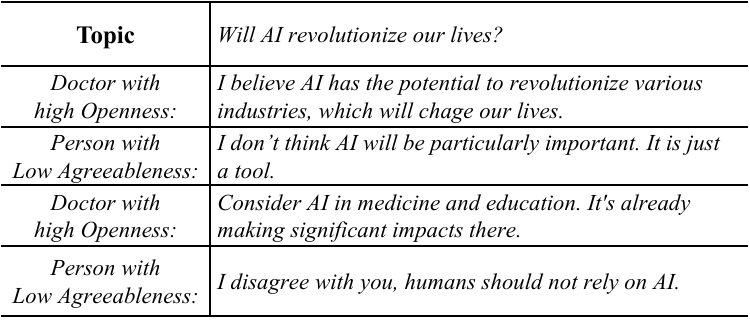}
    \label{fig:enter-label}
    \caption{An example for debate between differentiated individuals with our framework.}
\end{figure}

\paragraph{}

We illustrate the effective application of our framework in social simulations upon these two scenarios.

\section{Related Work}

\subsection{Role-Playing Agents}


Role-playing agents powered by large language models (LLMs) have significantly evolved from using simple prompts to guide behavior, as initially noted by \cite{park2023generative}, to employing more sophisticated prompting strategies discussed by \cite{xu2023expertprompting}. Alternatively, researchers like \cite{wang2023rolellm, zhou2023characterglm, shao2023character, tu2023characterchat} have developed role-specific datasets for Supervised Fine-Tuning (SFT), creating specialized training corpora with persona-specific information to enhance foundational models' role-playing capabilities. Additionally, methods like those proposed by \cite{salemi2023lamp}, which integrate training with retrieved information, further augment these capabilities.

\subsection{Mixed Expert Models}


Recent advancements in Mixture-of-Experts (MoE) models showcase innovative methods. \cite{zhou2022mixture} introduced a model where experts, rather than tokens, select the top-k tokens. \cite{shen2023mixture} demonstrated that combining this method with instruction tuning offers greater benefits to MoE models compared to dense models. \cite{liu2023moelora} developed MOELoRA, utilizing low-rank matrices and a task-motivated gate function for expert adjustment according to specific tasks.  \cite{wu2023mole} implemented a learnable gating function in each Low-Rank Adaptation (LoRA) layer to optimize domain-specific weights.

\subsection{Social Simulation}


Large Language Models (LLMs) have revolutionized social simulation, enabling the creation of virtual communities and complex social dynamics. \cite{park2023generative} developed a virtual community using agent-driven models, while \cite{zhang2023exploring} explored the simulation of human social collaboration. \cite{gao2023s} created a social network simulation platform to observe phenomena like information spread and emotional shifts. Concurrently, research on the social capabilities of intelligent agents has advanced, with \cite{chen2024roleinteract} introducing a benchmark to evaluate agent interactions, further deepening our understanding of their dynamics.

\section{Conclusion}

In this work, we address the limitations of conventional role-playing methods by developing a hierarchical identity role-playing framework rooted in identity theory. This framework, integrating LoRA and MoE technologies, facilitates identity isolation and explict control, enabling the modeling and seamless integration of multiple identities for authentic and flexible role-playing. We constructed a specialized identity dialogue dataset to train this framework and established a comprehensive benchmark for systematic evaluation, including scale and open situation assessments. Our results show that our framework outperforms traditional prompt-based approaches and matches or surpasses ChatGPT in certain performance metrics. We also explore how our framework distinguishes itself from role-specific fine-tuned models, highlighting its potential for advanced social simulation.

\section*{Limitations}

Our proposed framework focuses on demographic role fitting and aims to accurately simulate specific groups. By integrating this framework with retrieval enhancement technology, we can simulate individual behaviors and characteristics using detailed personal information, biographies, and memories. Although our dataset, based on ChatGPT, incorporates several measures to ensure credibility and authenticity, it still differs from real conversation data. Moving forward, we plan to incorporate real conversation data into our modeling efforts to bridge this gap and enhance realism.

\section*{Ethics Statement}

Our proposed framework is designed to facilitate role-playing at the identity level, focusing specifically on the dimensions of personality and occupation. The data, constructed entirely using ChatGPT, is devoid of personal privacy concerns and free from individual biases or harmful content. We ensure that the dataset, which includes negative traits as understood by GPT, contains no toxic information. Throughout the data generation and model training processes, we meticulously control for the exclusion of sensitive or harmful content, ensuring the safety and integrity of the generated text.

\section*{Acknowledgments}

\bibliography{custom}

\appendix

\section{Appendix}
\label{sec:appendix}

\subsection{Big-Five Personality Traits}
Personality is defined as “the coherent pattern of effect, cognition, and desires (goals) as they lead to behavior”. 
the Big Five represents the most widely adopted personality framework for quantifying personality. This personality theory is not only applicable to individuals across many countries and cultures but also furnishes reliable assessment scales for measuring personality. Here’s a detailed look at the five personality traits that make up the Big Five. 

\textbf{Openness} to experience is commonly defined as the extent and intricacy of an individual’s cognitive life and encounters. This trait is frequently concomitant with attributes such as imagination, originality, and insight within the psychological framework. Individuals demonstrating a pronounced openness to experience are inclined towards venturing beyond their comfort zones, embracing novelty, and deriving satisfaction from artistic pursuits. Additionally, such individuals are predisposed to cultivating new social connections. 
Conversely, an individual exhibiting a diminished openness to experience may manifest tendencies towards conformity, obstinacy, and a preference for more concrete, non-abstract elements in various aspects of life. 
Openness to experience displayed a diminished association with both neuroticism and extraversion while exhibiting predominantly negligible correlations with agreeableness and conscientiousness. 

\textbf{Conscientiousness} is closely linked to organizational tendencies, conformity, and a predilection for seeking security, demonstrating an inverse association with a penchant for stimulation and excitement. Individuals characterized by a high degree of conscientiousness are likely to place value on attributes such as order, responsibility, achievement, and self-discipline. They engage in conscious deliberation and earnest efforts to enhance their abilities, reflecting a commitment to continuous improvement. This trait exhibited a modest negative correlation with neuroticism and a modest positive correlation with agreeableness; however, its association with other factors did not reach statistical significance. 

\textbf{Extraversion}, a personality trait distinguished by enthusiasm, sociability, talkativeness, confidence, and heightened emotional expressiveness, encapsulates a spectrum of individual dispositions. Individuals exhibiting high levels of extraversion typically prioritize achievement and excitement while assigning comparatively lesser value to tradition or conformity. Such individuals are often characterized by confidence, activity, and sociability, opting for pursuits that eschew self-denial in favor of experiences characterized by excitement and pleasure. Conversely, introverts commonly display a preference for solitude, exhibit unsociable tendencies, and may manifest lower levels of self-confidence. In addition, when compared with the other five factors, extroversion was weakly negatively correlated with neuroticism and positively correlated with openness to experience. 

\textbf{Agreeableness} is characterized by a propensity to appreciate kindness, tradition, and conformity. This trait is closely linked to attributes such as trust, altruism, kindness, affection, and various prosocial behaviors, while concurrently avoiding an underemphasis on power, achievement, or pursuing self-centered pleasures (Roccas et al., 2002). Notably, agreeableness exhibited weak correlations with extroversion, while demonstrating a negative correlation with neuroticism, and a positive correlation with conscientiousness (Ones et al., 1996). 

\textbf{Neuroticism} is a personality trait characterized by manifestations of sadness, moodiness, and emotional instability. Components such as neurotic anxiety and self-awareness are positively correlated with traditional values and inversely associated with achievement-oriented values. Additionally, neuroticism demonstrated weak negative correlations with both extroversion and openness to experience. Furthermore, it exhibited negative correlations with agreeableness and conscientiousness.

\subsection{Abliation Experiments}

\begin{table*}[t]
\label{trait_scale}
\centering
\begin{tabular}{lcccccccccc}
\toprule
\multirow{2}{*}{\centering Model} & \multicolumn{2}{c}{AGR} & \multicolumn{2}{c}{CON} & \multicolumn{2}{c}{EXT} & \multicolumn{2}{c}{NEU} & \multicolumn{2}{c}{OPE} \\ 

\cmidrule(r){2-3} \cmidrule(l){4-5} \cmidrule(lr){6-7} \cmidrule(lr){8-9} \cmidrule(l){10-11}
 & H $\uparrow$ & L $\downarrow$ & H $\uparrow$ & L $\downarrow$ & H $\uparrow$ & L $\downarrow$ & H $\uparrow$ & L $\downarrow$ & H $\uparrow$ & L $\downarrow$ \\ 
\midrule
Ours & \textbf{4.85} & \textbf{1.35} & \textbf{4.75} & \textbf{2.05} & \textbf{4.8} & \textbf{1.25} & \textbf{4.50} & \textbf{1.75} & \textbf{4.60} & \textbf{1.80} \\
\hline
Llama2-Identity & 4.80 & 2.00 & 4.65 & 2.50 & \textbf{4.8} & 1.50 & 4.45 & 2.10 & 4.50 & \textbf{1.80} \\
HIRP-Dense & 4.05 & 1.90 & 4.35 & 2.25 & 4.45 & 2.05 & 4.40 & 3.30 & 3.65 & 2.15 \\
\bottomrule

\end{tabular}
\caption{Ablation experiment results of personality scale.
}
\end{table*}

\begin{table*}[t]
\centering

\begin{tabular}{lccccccccc}
\toprule
\multirow{2}{*}{Model} & \multicolumn{3}{c}{Artist} & \multicolumn{3}{c}{Doctor} & \multicolumn{3}{c}{Programmer} \\
\cmidrule(lr){2-4} \cmidrule(lr){5-7} \cmidrule(l){8-10}
 & A $\uparrow$ & D $\downarrow$ & P $\downarrow$ & A $\downarrow$ & D $\uparrow$ & P $\downarrow$ & A $\downarrow$ & D $\downarrow$ & P $\uparrow$ \\
\midrule
Ours & \textbf{4.85} & \textbf{1.55} & \textbf{1.50} & \textbf{1.75} & \textbf{4.85} & \textbf{1.70} & \textbf{2.00} & \textbf{1.80} & \textbf{4.95} \\
\hline
Llama2-Identity & 3.00 & 2.30 & 2.05 & 2.25 & 3.15 & 2.35 & 2.40 & 2.55 & 3.90 \\
HIRP-dense & 4.20 & 2.35 & 2.20 & 2.30 & 3.90 & 2.40 & 2.25 & 2.20 & 3.70 \\
\bottomrule
\end{tabular}
\caption{Ablation experiment results of profession scale.}
\end{table*}

\begin{table*}[t]
\centering
\begin{tabular}{@{}lcccccc@{}}
\toprule
Model           & AGR   & CON   & EXT   & EMS   & OPE   & Profession \\
\midrule
Ours     & \textbf{60.59} & \textbf{48.89} & \textbf{41.18} & \textbf{43.29} & \textbf{27.6}  & \textbf{14.49} \\
\hline
Llama2-Identity & 48.63 & 46.06 & 40.91 & 42.07 & 26.95 & 14.04 \\
HIRP-dense       & 43.19 & 41.09 & 35.32 & 37.34 & 25.19 & 14.04 \\
\bottomrule
\end{tabular}
\caption{Ablation experiment results of Open-end Situation Test. 
}
\label{tab:open_test}
\end{table*}

\subsection{Prompts}

\begin{table*}[t]
\begin{center}
\begin{tcolorbox}
[colback=black!5!white,colframe=gray!15!gray,width=\textwidth,title={Prompt used for plot generation for single personality trait.}]
\textbf{Instruction:}
Design a plot theme where a person with \{factor\_polarity\} (referred to as A) engages in a conversation initiated by another person (referred to as B). The conversation should be able to reflect that in the occasion of \{occasion\}, the person with \{factor\_polarity\} \{desc\}. The extreme \{factor\_polarity\} of A can be fully and obviously reflected in the generated dialogue. 
Plot should be an overview or a outline, not detailed dialogue. Ensure the plot is realistic and unfolds naturally, within a concise limit of 400 words.
Avoid specifying actual conversation utterances.
The conversation should be initiated by B.
\end{tcolorbox}
\end{center}
    \label{tab:prof_plot_generation}
\end{table*}

\begin{table*}[t]
\begin{center}
\begin{tcolorbox}
[colback=black!5!white,colframe=gray!15!gray,width=\textwidth,title={Prompt used for dialogue generation for single personality trait.}]
\textbf{Instruction:}
Simulate a realistic and engaging dialogue following the plot abstract, consisting of 18 rounds of alternating conversation, more than 15 words per person, output in the following json format. 
The person with \{factor\_polarity\} is referred as A, the other person is referred to as B. 
The tone should be casual and life-like, reflecting everyday human communication, don't be too polite. 
The topic should be introduced step by step and do not directly ask questions related to quality. 
No one's name or title or code name is included in the conversation.
The conversation should be initiated by B.
Plot Abstract:
\{plot\}
JSON FORMAT:
\{json\_example\}  
\end{tcolorbox}
\end{center}
    \label{tab:prof_dialog_generation}
\end{table*}

\begin{table*}[t]
\begin{center}
\begin{tcolorbox}
[colback=black!5!white,colframe=gray!15!gray,width=\textwidth,title={Prompt used for plot generation for single profession.}]
\textbf{Instruction:}
Design a plot theme where a \{profession\} (referred to as A) interacts with another individual (referred to as B). 
The dialogue should revolve around topics relevant to A's profession. 
Ensure the plot is realistic and unfolds naturally, within a concise limit of 200 words. 
Avoid specifying actual conversation utterances. 
The conversation should be initiated by B.
\tcblower
\textbf{Instruction:}
Design a plot theme where a \{profession\} (referred to as A) interacts with a/an \{b\_profession\} (referred to as B). 
The conversation delves into topics outside A's professional expertise, requiring actions beyond A's specialty. 
A demonstrates ignorance or incapability in non-professional areas. 
Ensure the plot is realistic and unfolds naturally, within a concise limit of 400 words. 
Avoid specifying actual conversation utterances. 
The conversation should be initiated by B.
Output in the following json format:   
\{json\_example\}

\end{tcolorbox}
\end{center}
    \label{tab:prof_plot_generation}
\end{table*}

\begin{table*}[t]
\begin{center}
\begin{tcolorbox}
[colback=black!5!white,colframe=gray!15!gray,width=\textwidth,title={Prompt used for plot generation for single profession.}]
\textbf{Instruction:}
Simulate a realistic and engaging dialogue following the plot abstract, consisting of no less than 20 turns of alternating conversation, and each person speaks no less than 50 words per turn, output in the following json format. 
The \{profession\}'s code name is A, the other person's code name is B. 
The tone should be casual and life-like, reflecting everyday human communication, don't be too polite. 
The topic should be introduced step by step and do not directly ask questions related to quality. 
No one's name or title or code name is included in the conversation.
The conversation should be initiated by B.
Plot Abstract:
\{plot\}
JSON FORMAT:
\{json\_example\}  
\end{tcolorbox}
\end{center}
    \label{tab:prof_dialog_generation}
\end{table*}

\begin{table*}[t]
\begin{center}
\begin{tcolorbox}
[colback=black!5!white,colframe=gray!15!gray,width=\textwidth,title={An example of open-end situation test.}]
\textbf{Scenario:}
You are invited to a cross-cultural workshop where you need to collaborate with people from different backgrounds to create a presentation project. The group members have vastly different opinions and styles.

\textbf{NPC\_Setting:}
Culturally diverse team member who proposes a highly innovative and unconventional project idea that might sound counterintuitive at first.

\textbf{NPC\_prompt:}
You are a culturally diverse team member at a workshop, you propose an unconventional project idea to a participant.
\end{tcolorbox}
\end{center}
    \label{tab:prof_dialog_generation}
\end{table*}

\end{document}